%% file: main.tex
\documentclass[sigconf,nonacm,screen,balance=false,10pt]{acmart}

\usepackage{xcolor}
\definecolor{revisionblue}{RGB}{0,76,180}

\definecolor{revisionblue}{RGB}{0,90,180}
\usepackage{pifont}
\usepackage{xurl}
\usepackage{booktabs}
\usepackage{tabularx}
\usepackage{array}
\usepackage[ruled]{algorithm2e}
\usepackage{algpseudocode}
\usepackage{enumerate}
\usepackage{enumitem}
\usepackage[english]{babel}
\usepackage{blindtext}
\usepackage{subfig}
\usepackage{amsmath}

\usepackage{amssymb}
\usepackage{xspace}
\usepackage{multirow}
\usepackage{threeparttable}
\usepackage{soul}
\usepackage{float}
\usepackage{flafter}
\usepackage{comment}
\usepackage{graphicx}
\usepackage{caption}

\usepackage{textcomp}

\def\fig{Fig.\xspace}

\newcommand{\head}[1]{{\noindent \textbf{#1:}}}

\graphicspath{{./fig/}}
\DeclareGraphicsExtensions{.pdf,.jpeg,.png}

\newcommand{\com}[1]{}
\newcommand{\todo}[1]{}

\setcopyright{none}
\date{}
\def\sysname{\textsc{StickFigure}\xspace}

\newcommand{\finalPeakActivationBytes}{254{,}608~B\xspace}
\newcommand{\finalPeakActivationKiB}{248.64~KiB\xspace}

\newcommand{\sfPacketBytes}{76~B\xspace}

\newcommand{\canonicalExpansionSchedule}{$(3,3,4,5)$\xspace}

\newcommand{\canonicalScheduleParams}{284.2K\xspace}
\newcommand{\canonicalScheduleMACs}{29.54M\xspace}

\AtBeginDocument{%
  \hypersetup{%
    pdfauthor={Aveline Xiaoyuan Han, Chenshu Wu},
    pdftitle={StickFigure: Enabling Pose-Only IoT Cameras for Privacy-Preserving Edge Vision}%
  }%
}

\begin{document}

\title[StickFigure]{\sysname: Enabling Pose-Only IoT Cameras for Privacy-Preserving Edge Vision}

\author{Aveline Xiaoyuan Han}
\affiliation{%
  \institution{The University of Hong Kong}
  \city{}
  \country{}
}
\email{xiaoyuanh@connect.hku.hk}

\author{Chenshu Wu}
\affiliation{%
  \institution{The University of Hong Kong}
  \city{}
  \country{}
}
\email{chenshu@cs.hku.hk}
\renewcommand{\shortauthors}{Han and Wu}

\begin{abstract}
Indoor IoT cameras can support assisted living and activity-aware spaces, but exporting raw RGB frames raises privacy concerns by exposing facial, clothing, and household details that body-motion tasks often do not require. We present \sysname, a pose-only IoT camera system that, during real-time monitoring, keeps raw frames local and ephemeral while exporting only compact skeleton streams sufficient for downstream human activity recognition (HAR).
While pose estimation has matured on GPUs, extracting accurate pose streams on resource-constrained IoT cameras remains difficult. To overcome these challenges and enable on-device pose streams on (almost) any IoT camera, we present two core designs in \sysname: an activation-aware pose backbone that allocates channel capacity after spatial downsampling and a lightweight hierarchical MLP refiner that selectively corrects elbow and wrist coordinates using intermediate visual features and kinematic context.
We fully implement \sysname and evaluate it on public benchmarks and real-world deployment.
Compared with state-of-the-art MCU pose models, \sysname achieves the highest pose accuracy and downstream HAR utility on both NTU RGB+D and Penn Action, while using 96.5\% fewer parameters and 99.3\% fewer MACs per frame than RTMPose-S-based baselines, achieving 0.107~s p95 latency on a representative STM32H753 MCU.
We believe \sysname establishes a new operating point for privacy-aware edge vision on massive IoT cameras.

\end{abstract}

\maketitle
\hypersetup{pdfauthor={Aveline Xiaoyuan Han, Chenshu Wu}}

\input{body}

\bibliographystyle{unsrt}
\bibliography{refs}

\appendix
\section{Fine-Arm Activity Subsets}
\label{app:fine-arm}

Fine-arm recall is the macro-average of per-class recall over the following predefined arm-centric subsets.

\noindent\textbf{NTU RGB+D 60:}
A010 clapping, A023 hand waving, A031 pointing, A033 check watch, A038 salute, A040 cross hands, A044 touch head, A046 touch back, and A049 use fan.

\noindent\textbf{Penn Action:}
Baseball pitch, baseball swing, bowling, golf swing, strumming guitar, tennis forehand, and tennis serve.

\end{document}

%% file: body.tex
\begin{figure}[t!]
  \centering
    \captionsetup{labelfont=bf,textfont=normalfont}
  \includegraphics[width=\columnwidth, trim={0 30pt 0 0},
  clip]{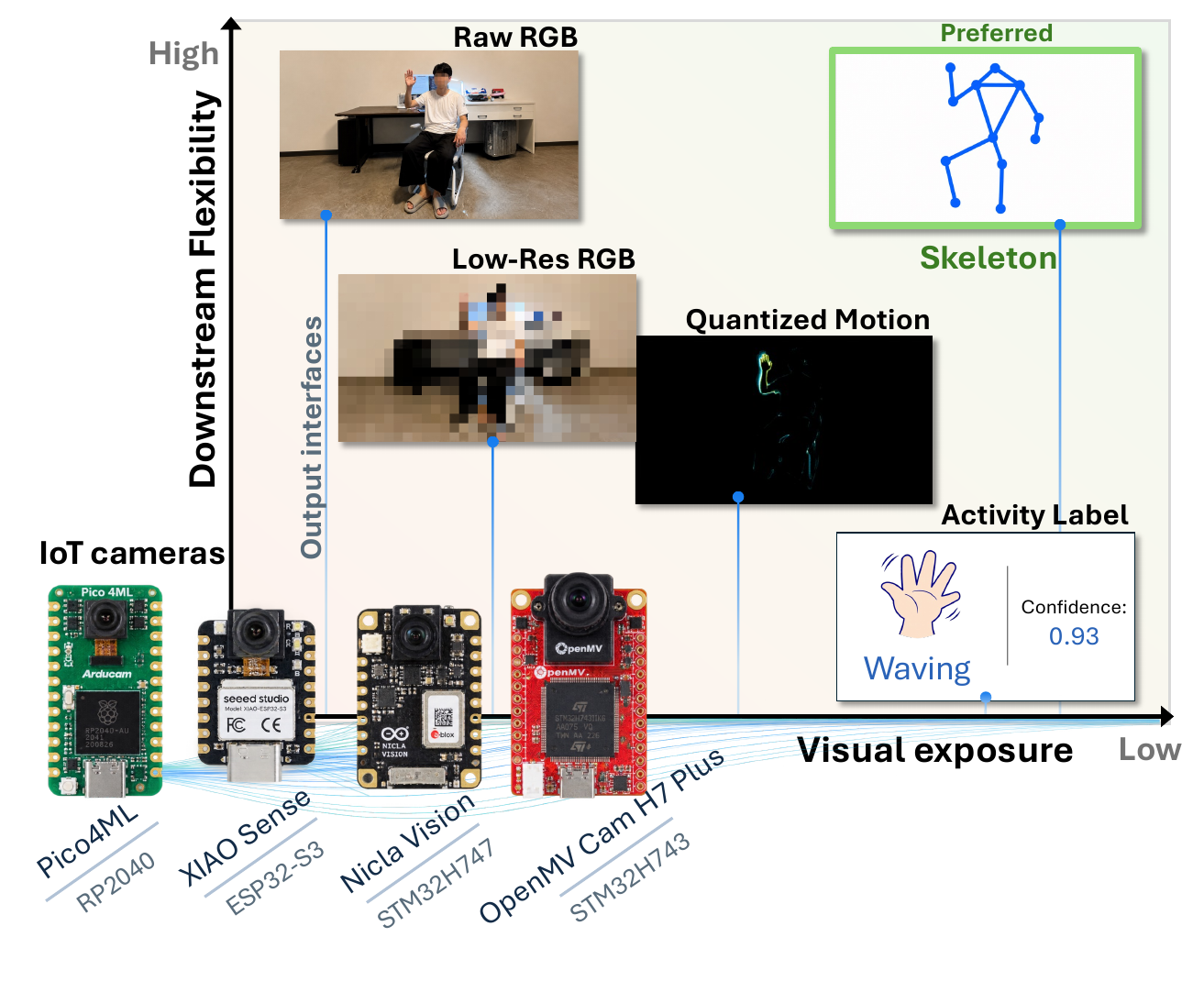}
  \caption{\textbf{Representation trade-offs for body-centric sensing.}
Qualitative comparison of visual exposure (lower toward the right) and downstream flexibility (higher toward the top). Pose reduces appearance-rich exposure while retaining a flexible body-motion representation.}
  \label{fig:interface_comparison}
\end{figure}

\section{Introduction}
\label{sec:introduction}

Internet-connected cameras are becoming common fixtures in private homes. In the United States alone, an estimated 74.9 million households, or 61\% of all households, had at least one indoor or outdoor security camera by 2026~\cite{safehome2026market}.
Beyond security, indoor cameras can support assisted living and activity-aware spaces through continuous body-motion sensing~\cite{bouchabou2021survey,sun2023human}. Continuous visual sensing in private spaces, however, raises privacy concerns~\cite{singh2018users,hussain2019different,chen2012sensor}. The reason is that most video-first activity-sensing pipelines transmit or retain raw RGB for analysis at a home hub, edge server, or in the cloud, exposing sensitive information including faces, clothing, and household context~\cite{beugin2022smartcamera,poddar2020visor,cangialosi2022privid,jana2013darkly,hinojosa2021privacyoptics,hinojosa2022privhar,shalawadi2024privacycovers}. As a side effect, retaining this bulky visual stream to the cloud also incurs communication and storage costs~\cite{googleNestBandwidth}.

Prior work attempts to reduce visual exposure at three levels: data access, visual capture, and output representation.
Data-access mechanisms confine image-rich data via access control, trusted execution, and mediated queries~\cite{beugin2022smartcamera,poddar2020visor,cangialosi2022privid,jana2013darkly,aditya2016ipic}.
Capture-level approaches manipulate optics, resolution, or hardware to limit acquired details~\cite{hinojosa2021privacyoptics,hinojosa2022privhar,ryoo2017lowresolution,kumawat2022motion,shalawadi2024privacycovers}, but they often still produce image-bearing outputs or require specialized hardware modifications.
At the output level, systems replace video with either inflexible, task-specific semantic labels~\cite{giordano2020batteryfree,lee2025hypercam} or implicit intermediate representations~\cite{mireshghallah2020shredder}. While intermediate representations retain downstream utility, they remain opaque to users and vulnerable to reconstruction or attribute-leakage attacks~\cite{lu2022preva,song2020informationleakage}.
Even where individual mechanisms offer formal guarantees or attack-specific privacy metrics, they do not provide a transparent, well-defined privacy level for the released data itself. In summary, existing systems lack an inspectable privacy-aware output interface that is semantically clear to users while retaining broad utility for downstream body-motion tasks, and few of them address MCU deployability for IoT cameras.

Our key insight is that \textit{most human-centric HAR tasks can be supported by articulated joint geometry and temporal body motion without retaining full appearance-rich RGB}~\cite{shi2019directedgnn,zhou2023discriminative}. Moreover, RGB action models may rely more on background than human motion, creating spurious shortcuts that degrade cross-environment generalization~\cite{chung2022backgroundbias}. Raw RGB can therefore expose unnecessary sensitive appearance while introducing task-irrelevant visual noise.
Built upon this observation, we select pose as the camera's output interface. Pose removes image-bearing content from the released interface without requiring specialized optical or physical capture hardware. Its predefined joint semantics make the released information privacy-inspectable (unlike an opaque embedding), while remaining reusable across body-motion recognizers (unlike a fixed activity label). Skeletons preserve articulated body geometry and temporal motion while omitting much of the facial, clothing, background, and scene appearance present in RGB~\cite{shi2019directedgnn,zhou2023discriminative}. As \fig~\ref{fig:interface_comparison} illustrates, pose reduces visual exposure relative to RGB while retaining greater downstream flexibility than a fixed activity label. Its compact fixed-format representation also reduces communication and storage relative to video. We therefore ask how task-useful pose extraction can be sustained in real time across heterogeneous IoT-camera resource tiers, down to commodity MCUs.

At first glance, pose estimation may appear to be a solved vision problem: modern estimators achieve high accuracy on GPUs and even mobile processors. A camera-class MCU, however, exposes three coupled challenges: severe resource constraints, uneven fine-joint degradation under compression, and variability across deployment sites and hardware tiers. Sustaining a real-time, task-useful pose stream under all three remains an open problem for complete IoT cameras.

\head{Severe MCU Resource Constraints} Commodity MCU-class IoT cameras offer only a few megabytes of Flash and a few hundred kilobytes to about 1 MB of SRAM (e.g., nRF52840 and STM32H753)~\cite{stm32h753_st, nrf52840_nordic}, whereas lightweight pose models still contain millions of parameters and require hundreds of millions to billions of operations per frame (e.g., YOLO26n-Pose and RTMPose-t/s)~\cite{ultralytics_yolo_pose,jiang2023rtmpose}.
For example, the 2.9M parameters of YOLO26n-Pose alone require about 2.9~MB in INT8, exceeding the target's Flash before firmware is included. Intermediate feature maps must also share limited SRAM with camera buffers, runtime workspace, communication state, and firmware. In particular, channel expansion at high spatial resolution can create activation tensors that exceed the memory available to the camera pipeline, even when parameter count appears modest. The challenge is therefore to achieve sufficient pose accuracy while simultaneously satisfying the storage and computation constraints of an MCU-only camera and limiting large intermediate activation tensors~\cite{lin2020mcunet,lin2021mcunetv2}.

\head{Uneven Joint Degradation under Compression} Aggressive model compression does not affect all joints equally. Compact pose models can preserve coarse body structure while leaving larger localization errors at elbows and wrists. These joints carry motion that is particularly important for arm-centric activities, while the commonly used aggregate percentage of correct keypoints (PCK) can mask such joint-specific degradation.
Global increases in backbone capacity can improve accuracy but impose additional computation across the entire pose estimator. The challenge is therefore to preserve task-critical distal-joint fidelity under aggressive compression without globally increasing model capacity and eroding the resource savings required for MCU deployment.

\head{Deployment Variability and Resource Heterogeneity} Dataset accuracy alone does not establish whether the system remains useful after deployment. New camera sites introduce visual shift, inactive periods make continuous pose inference wasteful, and different IoT camera tiers impose different resource budgets. A practical pose-only camera must therefore tolerate site-specific visual variation, avoid unnecessary computation during long inactive periods, and remain useful across devices with substantially different memory and compute budgets. This requires evaluating pose accuracy and downstream HAR as the system moves from benchmark data to live environments and heterogeneous hardware.

\begin{figure*}[t!]
  \centering
    \captionsetup{labelfont=bf,textfont=normalfont}
  \includegraphics[width=\textwidth]{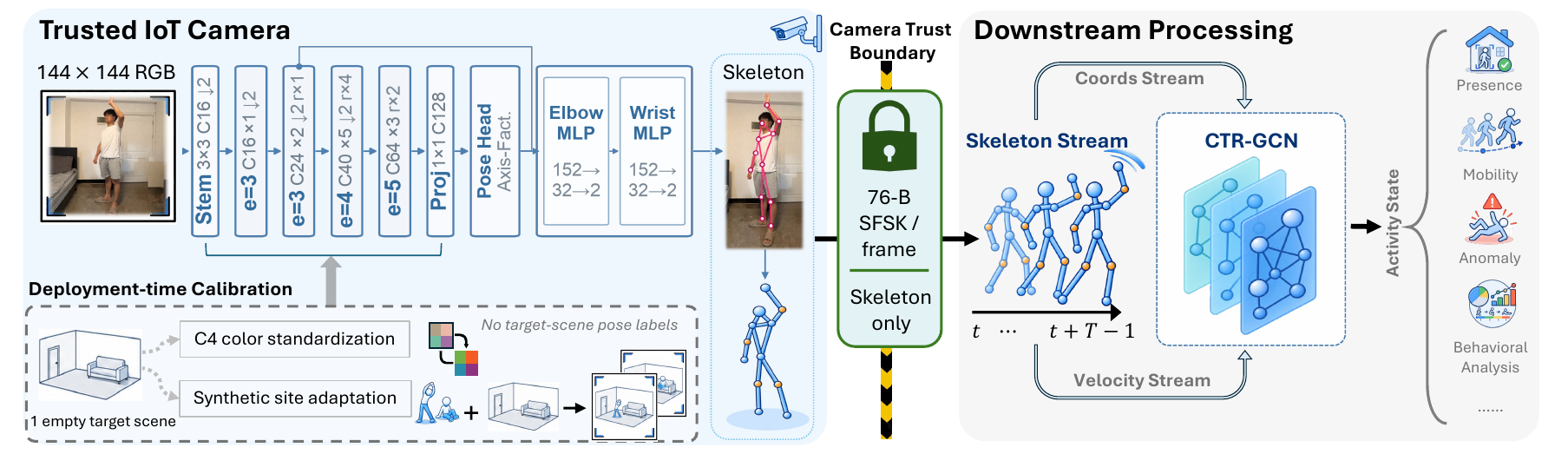}
  \caption{\textbf{\sysname system overview.} During real-time monitoring, RGB remains within the trusted camera boundary. An activation-aware pose backbone and a hierarchical MLP refiner produce a 12-joint pose, which is exported as a \sfPacketBytes fixed-format skeleton record (SFSK) for downstream HAR. The dashed path shows one-shot site calibration.}
  \label{fig:overview}
\end{figure*}

As \fig~\ref{fig:overview} outlines, \sysname addresses these challenges through an integrated system--model co-design that combines memory-efficient pose estimation, targeted refinement of compression-sensitive joints, and system-level optimization for site robustness and cross-device scalability. Specifically, we first propose an activation-aware pose backbone that keeps high-resolution stages narrow and shifts channel capacity to later, downsampled stages, improving capacity allocation under tight MCU memory and compute budgets. Rather than increasing capacity throughout the estimator, we employ a lightweight hierarchical MLP to combine intermediate visual features with kinematic context to selectively refine elbows and wrists. To bridge the deployment domain gap, we introduce one-shot, target-free site calibration: a single empty-scene snapshot from the target site is sufficient to construct scene-conditioned training data and adapt the pose estimator to the deployment environment, without observing or labeling the target user. Motion-triggered adaptive execution further reduces power during inactive periods.

We prototype the complete \sysname camera pipeline on an STM32H753 MCU with a 480-MHz Arm Cortex-M7 core and an OV2640 image sensor. The prototype integrates image capture, preprocessing, INT8 pose inference, distal-joint refinement, skeleton encoding, and communication.
Our evaluation spans two public benchmarks, three indoor deployment sites, a separate long-horizon single-person case study with an approximately 40-minute annotated session, and four hardware platforms across ESP32-S3, STM32H753, Raspberry Pi 4, and Jetson Orin Nano.
Our results show that, among MCU-scale methods, \sysname achieves the highest pose accuracy and downstream HAR utility on both benchmarks. Relative to XiNet Pose, it improves PCK by 8.8--11.8\%. It also uses 51.6\% fewer parameters, 52.1\% less Flash, 60.6\% fewer MACs, and 12.3\% less peak activation RAM. Compared with the strong RTMPose-S-based reference for more capable edge hardware, \sysname reduces parameter count by 96.5\% and MACs by 99.3\%, with relative PCK and HAR Top-1 decreases of 8.4--17.5\% and 8.0--17.0\%, respectively.
\sysname runs in real time on STM32H753 MCU, with a 0.107~s end-to-end latency per frame, and achieves 96.45\% macro-F1 (compared with 89.31\% for XiNet Pose) for HAR in a real-world deployment.
Together, these results demonstrate that accurate and reusable pose streams can be sustained efficiently across highly resource-constrained IoT cameras without exporting RGB frames, underpinning large-scale adoption on any IoT cameras for ubiquitous edge vision with privacy.

\head{Contributions} Our core contributions are as follows:

\par\noindent$\bullet$\ We present \sysname, an end-to-end pose-only IoT camera system that performs continuous pose extraction on MCU-class hardware while keeping RGB local and exporting only compact skeleton streams.

\par\noindent$\bullet$\ We design an activation-aware pose estimator that shifts channel capacity to downsampled stages and selectively refines compression-sensitive joints, preserving distal-joint fidelity under tight MCU memory and compute budgets.

\par\noindent$\bullet$\ We implement \sysname as an integrated real-time STM32H753 camera system with one-shot, target-person-free site calibration and motion-triggered execution, and extensively validate its performance across public benchmarks, realistic indoor deployments, and four hardware tiers.

\section{Motivation and Scope}
\label{sec:scope}

\noindent\textbf{Pose interface and joint selection.}
For body-centric sensing, pose preserves articulated body geometry and temporal motion while suppressing much of the facial, clothing, background, and scene appearance present in RGB~\cite{shi2019directedgnn,zhou2023discriminative}. Its predefined joint semantics also align naturally with graph-based HAR models operating on joint trajectories and joint--bone relations~\cite{yan2018stgcn,shi2019twostream,chen2021ctrgcn,sun2023human}. Unlike a fixed activity label, a skeleton stream does not bind the camera to one recognizer or activity vocabulary~\cite{giordano2020batteryfree,lee2025hypercam}. Unlike an implicit learned representation, its coordinates retain directly interpretable physical meanings rather than exposing an opaque feature space~\cite{song2020informationleakage}. Pose therefore provides a middle-ground interface: it is narrower than video, more inspectable than learned features, and applicable to a broader range of body-motion tasks than a fixed semantic output. We instantiate this interface as a 12-joint skeleton containing the head, a torso/root joint, and bilateral shoulders, elbows, wrists, knees, and ankles. Together, these joints capture the posture and motion patterns needed for the assisted-living activities targeted in this work. We treat this schema as a compact interface choice rather than a universally optimal joint configuration.

\noindent\textbf{Privacy scope.}
\sysname targets private indoor environments where continuous body-motion sensing is useful but exporting appearance-rich video is undesirable, including bedrooms, living spaces, and assisted-living environments. Specifically, private rooms can capture people while sleeping, dressing, or receiving care, exposing faces, expressions, clothing, and states of undress rarely visible in public. This concern directly shapes user behavior: some residents exclude sensing technologies from bedrooms, while acceptance of video-based assisted living falls sharply for undressing and other intimate activities~\cite{burrows2018privacy,maidhof2023eyes}. To limit this exposure, we place the trusted boundary at the camera MCU, the earliest software-controlled stage after image capture. The MCU and trusted firmware perform pose extraction and discard each RGB frame before any monitoring image leaves the camera. This minimizes the components and communication links exposed to appearance-rich data. The communication path, downstream applications, and stored skeleton histories remain outside the trusted boundary and therefore may observe the exported stream. Thus, the released representation exposes joint geometry and motion rather than monitoring RGB containing faces, expressions, clothing, or other pixel-level appearance. Beyond this privacy boundary, the current system focuses on single-person or primarily single-person activity. Applications requiring faces, text, object identity, fine hand--object state, or detailed scene context fall outside the scope of this pose interface.

\noindent\textbf{MCU target.}
The MCU families represented in \fig~\ref{fig:interface_comparison} span a broad range of on-chip compute and memory resources. The STM32H753 lies at the high-resource end of this MCU-camera spectrum, making it a practical primary platform for validating the complete camera pipeline.
It nevertheless remains substantially more constrained than SBC- and accelerator-class platforms, providing only a 480~MHz Arm Cortex-M7 core, 2~MB of Flash, and 1~MB of SRAM~\cite{stm32h753_st}.
Because this SRAM must be shared among camera buffers, preprocessing, runtime workspace, communication state, firmware, and model tensors, we impose a 256~KiB architecture-level limit on the largest individual activation tensor. The STM32H753 defines our primary operating point, while \S~\ref{sec:eval-cross-platform} evaluates the architecture across broader resource tiers.

\begin{figure}[t]
  \centering
    \captionsetup{labelfont=bf,textfont=normalfont}
  \includegraphics[width=\columnwidth]{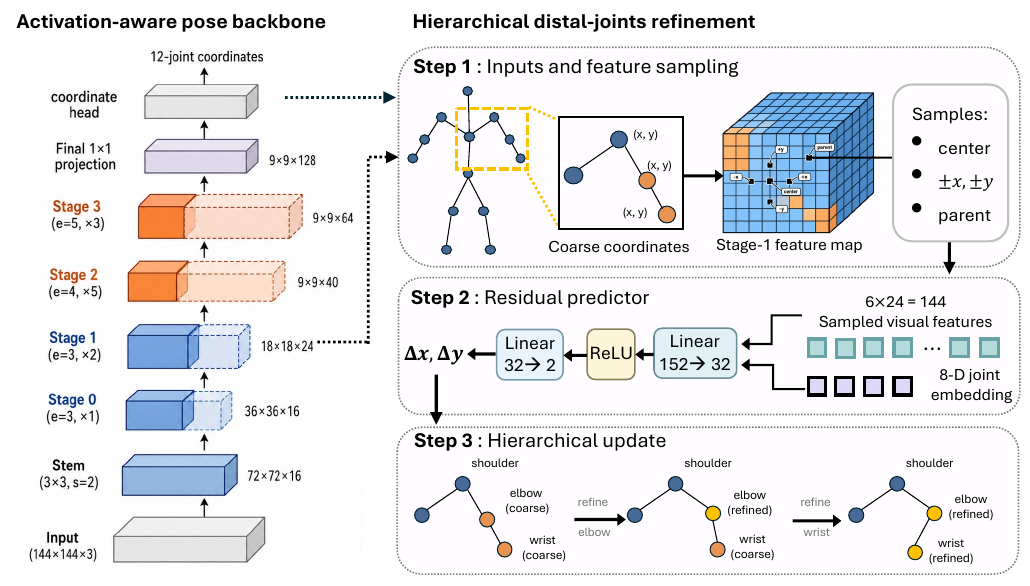}
  \caption{\textbf{Activation-aware pose backbone and hierarchical joint refinement.} The backbone delays larger channel expansion until after spatial downsampling, while the joint refiner uses intermediate features and kinematic-parent context to correct elbows and wrists.}
  \label{fig:pose-backbone-refiner}
\end{figure}

\section{System Design}
\label{sec:overview}

\subsection{\sysname Overview}
\fig~\ref{fig:overview} presents the end-to-end \sysname pipeline. The camera-side path estimates a 12-joint pose using an activation-aware backbone, selectively refines the distal arm joints with a hierarchical MLP, and packages the result as an SFSK record for skeleton-only export. Downstream software transforms the resulting skeleton sequence into temporal motion features for HAR. One-shot site calibration and motion-triggered execution complement this primary path by addressing deployment shift and inactive-period computation.

\subsection{Activation-Aware Pose Extraction}
\label{sec:activation-backbone}
The camera-side pose estimator comprises an inverted-bottleneck backbone and lightweight coordinate and confidence heads. Its central design question is how to distribute channel capacity across spatial scales without exceeding the MCU activation-memory budget. We first establish this allocation principle, then describe the backbone and prediction heads.
\subsubsection{Capacity Allocation}
MCU feasibility depends not only on parameter count and MACs, but also on the size of intermediate activation tensors that share SRAM with the camera and runtime. For an INT8 expanded tensor with height \(H\), width \(W\), and \(C_{\mathrm{exp}}\) channels, its size scales as
\begin{equation}
M_{\mathrm{act}} = HWC_{\mathrm{exp}} \text{ bytes}.
\label{eqn:activation-memory}
\end{equation}
 Because this cost scales with spatial area, channel expansion is most expensive at high resolution. The backbone therefore keeps high-resolution stages narrow and introduces larger expansion ratios only after spatial downsampling. A stride-2 reduction along both spatial axes reduces the activation area by a factor of four, making later channel expansion substantially less memory-intensive. This limits large high-resolution intermediate activation tensors while allowing parameters and computation to grow selectively in later stages. \fig~\ref{fig:pose-backbone-refiner} shows this allocation and joint-refinement path.

\subsubsection{Backbone Architecture}

Each pose pass receives a \(144\times144\times3\) input tensor. For a source frame of width \(W\) and height \(H\), preprocessing selects the centered square of side \(\min(W,H)\), converts it to RGB, resizes it bilinearly, and applies the model normalization. The model predicts normalized joint coordinates relative to the cropped field of view, \((x,y)\in[0,1]^2\). When needed, the inverse crop-and-resize transform maps the predictions back to the original frame.

Following the allocation principle above, the backbone uses a four-stage inverted-bottleneck architecture that progressively reduces spatial resolution while increasing channel capacity. A \(3\times3\), stride-2 convolution followed by BatchNorm and ReLU6 maps 3 channels to 16 and produces a \(72\times72\) feature map. Four inverted-bottleneck stages use the canonical expansion schedule \canonicalExpansionSchedule, with one ratio applied to every block in each stage. Stage~0 uses one ratio-3 block to downsample to \(36\times36\) while retaining 16 output channels. Stage~1 uses two ratio-3 blocks to reach \(18\times18\) and 24 channels. Stage~2 uses five ratio-4 blocks to reach \(9\times9\) and 40 channels. Stage~3 uses three ratio-5, stride-1 blocks at \(9\times9\) and produces 64 channels. A final \(1\times1\) convolution, BatchNorm, and ReLU6 projects the feature map to \(128\times9\times9\). Each downsampling stage applies stride 2 only in its first block; later blocks use stride 1. Residual connections are enabled only when input and output shapes match. Each inverted bottleneck contains a \(1\times1\) expansion convolution, a \(3\times3\) depthwise convolution, and a \(1\times1\) projection convolution, with BatchNorm after every convolution and ReLU6 after expansion and depthwise filtering. We omit squeeze-and-excitation and attention to retain a compact operator set compatible with MCU INT8 execution.

\subsubsection{Coordinate and Confidence Heads} The coordinate head predicts joint locations directly rather than materializing dense heatmaps. Global average pooling converts the \(128\times9\times9\) feature map to a 128-dimensional vector, followed by a 64-dimensional linear layer and ReLU. Independent horizontal and vertical heads each produce \(12\times48\) logits. A softmax over the 48 bins yields an axis-wise distribution, and its expectation over normalized bin centers gives the coarse coordinate. A separate 24-dimensional head predicts horizontal and vertical residuals for all 12 joints. A \(\tanh\) bounds each residual to half a bin width before addition, after which coordinates are clipped to \([0,1]\). This head preserves sub-bin correction without introducing heatmap-sized activation tensors. A separate confidence head outputs one value per joint, with shape \([B,12]\). The subsequent joint refiner updates only elbow and wrist coordinates, so the backbone confidence values are retained in the final pose.
The canonical backbone contains \canonicalScheduleParams parameters and uses approximately \canonicalScheduleMACs MACs per frame. Its profiled peak activation memory is \finalPeakActivationKiB. The backbone also provides its \(24\times18\times18\) Stage~1 feature map to the hierarchical joint refiner.

\subsection{Hierarchical Distal-Joint Refinement}
\label{sec:refiner}

Elbows and wrists are particularly sensitive to aggressive MCU compression. These joints carry discriminative motion for arm-centric activities, while globally increasing backbone capacity would add computation to every joint. We therefore use a lightweight, frame-local hierarchical MLP to refine only the left and right elbows and wrists.

\subsubsection{Feature Sampling} \fig~\ref{fig:pose-backbone-refiner} shows the feature-sampling and hierarchical update path. The refiner receives the normalized coarse pose \([B,12,2]\) and the backbone's Stage~1 feature map \([B,24,18,18]\), which retains more spatial detail than the final \(9\times9\) representation. For each refined joint, the module bilinearly samples six 24-dimensional feature vectors: one at the coarse joint location, four at neighboring offsets along the image axes, and one at its kinematic parent. Their concatenation produces a 144-dimensional visual descriptor. An 8-dimensional learned joint-type embedding is appended to form the 152-dimensional predictor input. The local stencil captures nearby visual evidence, while the parent sample introduces arm-chain context.

\subsubsection{Kinematic Update} The elbow and wrist predictors each apply \(\mathrm{Linear}(152,32)\), ReLU, and \(\mathrm{Linear}(32,2)\) to predict coordinate residuals \((\Delta x,\Delta y)\). The two elbows share one predictor and the two wrists share another, while elbow and wrist predictors use distinct weights. Refinement proceeds from proximal to distal. The elbow is first updated using the shoulder as its kinematic parent,
\begin{equation}
    \mathbf{p}'_e
    =
    \operatorname{clip}(\mathbf{p}_e+\Delta\mathbf{p}_e,0,1),
    \label{eqn:elbow-update}
\end{equation}
and the refined elbow then provides the parent context for the wrist update,
\begin{equation}
    \mathbf{p}'_w
    =
    \operatorname{clip}
    \left( \mathbf{p}_w+\Delta\mathbf{p}_w(\mathbf{p}'_e),0,1
    \right).
    \label{eqn:wrist-update}
\end{equation}
Each residual component is bounded to \([-0.10,0.10]\), and the updated coordinates are clipped to \([0,1]\). Only the four distal arm joints are modified; the remaining eight joints retain their backbone predictions. This proximal-to-distal update allows the corrected elbow to guide wrist refinement without introducing a second full-body pose head. The refinement path adds 10.1K parameters and 0.02M MACs per frame without increasing the profiled peak activation memory.

\subsection{Skeleton Streaming and HAR}
\label{sec:skeleton-har}

After joint refinement, each frame is encoded as a \sfPacketBytes fixed-format skeleton record (SFSK). Each record contains quantized joint coordinates, per-joint confidence and validity information, packet-ordering metadata, and an integrity check. The fixed record size bounds per-frame communication, while its ordering and validity fields allow downstream software to identify missing or unusable observations.

\subsubsection{Temporal Representation}
Downstream HAR operates on sequences of predicted skeletons rather than independent poses. Let \(\mathbf{p}_{t,j}\in[0,1]^2\) denote joint \(j\) at observation \(t\), and let \(r\) denote the torso/root joint. We construct root-centered coordinates
\begin{equation}
    \mathbf{q}_{t,j}
    =
    \mathbf{p}_{t,j}-\mathbf{p}_{t,r},
    \label{eqn:root-centered-position}
\end{equation}
velocity between consecutive valid observations
\begin{equation}
    \mathbf{v}_{t,j}
    =
    \frac{\mathbf{q}_{t,j}-\mathbf{q}_{t',j}}{\Delta t},
    \label{eqn:joint-velocity}
\end{equation}
where \(t'\) is the preceding valid observation and \(\Delta t\) accounts for frame gaps, and bone vectors
\begin{equation}
    \mathbf{b}_{t,j}
    =
    \mathbf{q}_{t,j}
    -
    \mathbf{q}_{t,\operatorname{parent}(j)}.
    \label{eqn:bone-vector}
\end{equation}
Concatenating the two coordinate dimensions of position, velocity, and bone features produces six channels. Each sequence is resampled to \(T=64\) observations and represented as \([N,6,64,12]\). Confidence and validity guide sequence construction but are not HAR input channels.

\subsubsection{Downstream HAR}
CTR-GCN~\cite{chen2021ctrgcn} processes these sequences on the corresponding 12-joint graph. For each pose estimator, we fine-tune a separate HAR model using skeleton sequences predicted by that estimator. The resulting HAR model is therefore exposed to the estimator's characteristic coordinate errors and temporal behavior. This pose-source-matched protocol measures the utility of the emitted skeleton stream rather than that of idealized ground-truth poses.

\subsection{One-Shot Site Calibration}
\label{sec:calibration}
Deployment sites can differ from benchmarks in background appearance, illumination, and camera color response. Collecting and annotating target-person poses per site would impose substantial setup overhead. \sysname instead adapts its pose estimator without observing the deployment user.

\subsubsection{Calibration Input and Synthesis}
As illustrated by the dashed path in \fig~\ref{fig:overview}, calibration uses a single empty-scene RGB image collected during setup, without target-person images or pose annotations. Labeled source-person foregrounds are composited onto the target background to construct site-conditioned training samples. The calibration image also provides statistics for camera and color standardization, while pose supervision comes from labeled source foregrounds.

\subsubsection{Bounded Model Adaptation}
Calibration starts from the pretrained pose model and runs offline before the updated model is installed on the camera. We update only the final backbone stage, feature projection, and coordinate and confidence heads. Earlier backbone stages, BatchNorm parameters, and the hierarchical joint refiner remain fixed. Target-scene composites are mixed with NTU replay samples to retain the original pose capability while adapting the trainable layers to the target scene. The adapted model is then frozen for real-time monitoring, so calibration adds no computation to the per-frame camera path.

\subsection{MCU Runtime}
\label{sec:mcu}

\subsubsection{Continuous Camera Pipeline}
The production implementation pairs an OV2640 camera with the STM32H753\allowbreak ZIT6 MCU. The OV2640 delivers \(144\times144\) RGB565 frames through DCMI/DMA into two alternating capture buffers. DMA fills one buffer while the previously completed buffer is available to software, decoupling sensor capture from inference. Before each pose pass, the MCU copies the newest completed safe frame into a dedicated processing buffer, preventing inference from reading a buffer that DMA may overwrite. When capture outpaces pose inference, a latest-frame-wins policy selects the most recent completed frame and skips older frames rather than building a queue. This keeps frame memory bounded and prioritizes fresh observations for the live skeleton stream. The MCU then performs RGB565 conversion and normalization, INT8 backbone inference, coordinate decoding, hierarchical joint refinement, SFSK construction, and UART transmission. Capture, processing, model-runtime, and packet buffers are statically allocated. During monitoring, each processed RGB frame is discarded after pose extraction, and the production path retains no RGB history. Together, static allocation and latest-frame-wins scheduling prevent frame queues or retained image histories from growing during continuous operation.

\subsubsection{Motion-Triggered Execution}
Continuous full-rate pose inference is unnecessary when the monitored scene is inactive. \sysname therefore operates in two runtime states: \texttt{MOTION\_CHECK} and \texttt{POSE\_INFERENCE}. In \texttt{MOTION\_CHECK}, the MCU performs a lightweight frame-difference check once per second and skips the pose path. Detected motion transitions the camera to \texttt{POSE\_INFERENCE} and restores the full pose path. After approximately 5~s without motion, the controller returns to \texttt{MOTION\_CHECK}. Only pose-path invocation changes between states; the model and SFSK record format remain unchanged.
We will evaluate the on-device power and latency gains from adaptive execution in \S\ref{sec:eval-live} and \S~\ref{sec:eval-overall}.

\section{Evaluation}
\label{sec:evaluation}

\subsection{Experimental Setup}
\label{sec:eval-setup}

We evaluate \sysname from model accuracy to real-world deployment. We first measure pose accuracy on two datasets and downstream HAR utility, then evaluate live deployment through one-shot calibration and motion-triggered adaptive execution. We further study long-horizon single-elderly monitoring as an end-to-end case study and examine scaling across heterogeneous edge platforms.

\subsubsection{Datasets}
\sysname is evaluated on NTU RGB+D 60~\cite{shahroudy2016ntu} and Penn Action~\cite{zhang2013actemes}. For NTU RGB+D 60, we use the official cross-subject protocol and select 20 pose-dominant activity classes spanning body transitions, arm-centric motions, and posture- and floor-related activities. The resulting subset contains 13,440 clips from 40 subjects and includes all three synchronized camera viewpoints. NTU serves as the primary benchmark for pose and downstream HAR evaluation. Penn Action contains 2,326 RGB video sequences covering 15 actions with frame-level 2D joint annotations. Its videos span diverse subjects, viewpoints, backgrounds, and capture conditions. We use the standard split of 1,258 training and 1,068 test sequences and fine-tune each pose estimator before evaluation. Penn Action tests whether the observed accuracy and resource trade-offs persist under dataset-specific training on a different action dataset.

\subsubsection{Baselines}
For controlled comparison, Tiny MobileNet and XiNet use the same pose head as our backbone variants to isolate backbone design. We denote them Tiny MobileNet Pose and XiNet Pose; the latter is an XiNet-based baseline, not a reproduction of the original XiNet-Pose system. We further compare Uniform-Expansion, Progressive-Expansion, our backbone alone, and full \sysname with hierarchical distal-joint refinement. All MCU-scale methods share dataset splits, training, and evaluation, so differences primarily reflect architecture. YOLOv8n-Pose and YOLO11n Detector + RTMPose-S provide higher-resource accuracy references.

\subsubsection{Evaluation Metrics} We use the following metrics:

\noindent\textbf{Pose accuracy.}
PCK@0.10 uses an axis-aligned bounding box over all valid ground-truth joints in each frame, with person scale \(s=\max(w_{\mathrm{bbox}},h_{\mathrm{bbox}},48~\mathrm{px})\). A prediction is correct if its Euclidean localization error is below \(0.10s\), and PCK is micro-averaged over valid joints. FP32 PCK is used for comparison with non-MCU accuracy references, while INT8 PCK reflects on-device accuracy.

\noindent\textbf{HAR utility.}
We report HAR Top-1 accuracy and fine-arm recall, the macro-average recall over predefined arm-centric activity subsets listed in Appendix~\ref{app:fine-arm}. CTR-GCN~\cite{chen2021ctrgcn} is fine-tuned separately for each pose estimator on predicted 64-frame skeleton sequences, so downstream evaluation reflects camera-side pose errors. HAR Top-1 is reported as the three-seed mean \(\pm\) sample standard deviation.

\noindent\textbf{Model resources.}
Model efficiency is characterized by parameter count, MACs per frame, Flash storage, and profiled peak activation RAM. Flash includes compiled INT8 weights and associated constants. Peak activation RAM is the maximum simultaneous footprint of live activation tensors during inference, excluding runtime code and application memory.

\noindent\textbf{Latency.}
We report the p95 end-to-end on-device pose latency from camera-frame readiness to skeleton readiness, including preprocessing, INT8 inference, and coordinate decoding. For \sysname, this also includes the refiner.

\subsubsection{Platforms} 
\begin{figure}[t]
\centering
\captionsetup{labelfont=bf,textfont=normalfont}
\includegraphics[width=0.95\columnwidth]{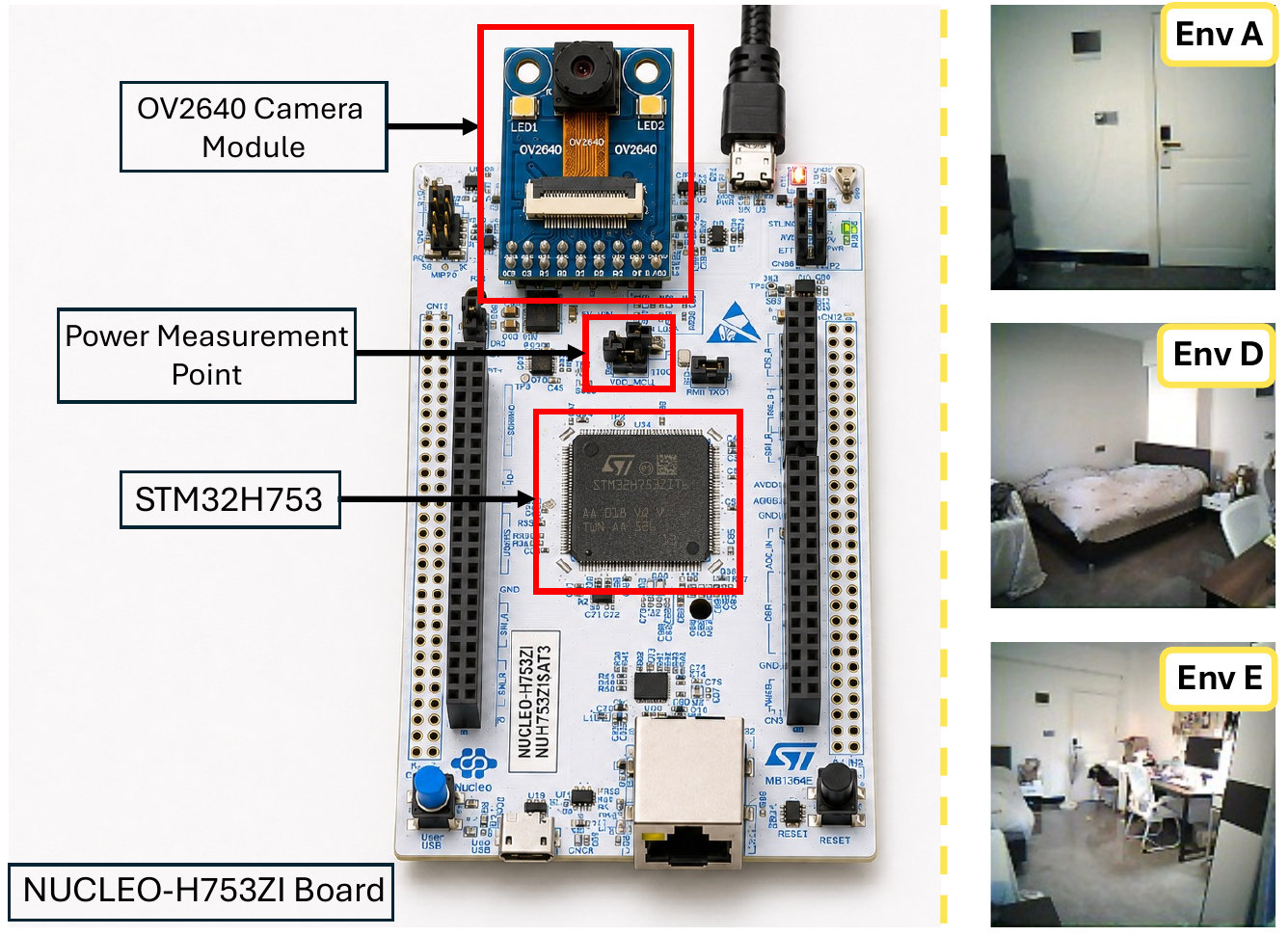}
\caption{\textbf{Hardware prototype and deployment environments.} \textbf{Left:} the NUCLEO-H753ZI prototype with an STM32H753\allowbreak ZIT6 MCU, an OV2640 camera module, and the board-level power-measurement point. \textbf{Right:} views of deployment Envs. A, D, and E used for live evaluation in \S~\ref{sec:eval-live}.}
\label{fig:hardware-platform}
\end{figure}

Our primary live-camera platform uses an STM32H753\allowbreak ZIT6 paired with an OV2640 capturing $144\times144$ RGB565 frames through DCMI/DMA. \fig~\ref{fig:hardware-platform} shows the NUCLEO-H753ZI prototype used for runtime, MCU-side power, energy, communication, and continuous-operation measurements. For cross-platform scaling, we additionally evaluate ESP32-S3, Raspberry Pi~4, and Jetson Orin Nano.

\subsection{Overall Performance}
\label{sec:eval-overall}

 Table~\ref{tab:end-to-end-comparison} compares FP32 PCK@0.10, p95 pose latency, HAR, fine-arm recall, and model resources across MCU-scale methods and non-MCU references. We examine the resulting operating points from three perspectives: pose and downstream utility, resource efficiency, and INT8 robustness.

\begin{table*}[t]
  \caption{End-to-end FP32 pose, p95 pose latency, HAR, and resource comparison on NTU RGB+D and Penn Action. Flash reports compiled INT8 weights and associated constants, while peak activation RAM reports the maximum simultaneously live model activation memory in the profiled NTU compiled graph. Non-MCU models are accuracy references; dashes indicate unavailable exact resource or target-latency measurements.}
  \label{tab:end-to-end-comparison}
  \centering
  \scriptsize
  \begingroup
  \renewcommand{\arraystretch}{1.08}
  \setlength{\tabcolsep}{1.7pt}
  \begin{tabularx}{\textwidth}{@{}>{\raggedright\arraybackslash}X*{12}{c}@{}}
    \toprule
    \multirow{2}{*}{\textbf{Method}} &
    \multicolumn{4}{c}{\textbf{Model resources}} &
    \multicolumn{4}{c}{\textbf{NTU RGB+D}} &
    \multicolumn{4}{c}{\textbf{Penn Action}} \\
    \cmidrule(lr){2-5}\cmidrule(lr){6-9}\cmidrule(l){10-13}
    & Params & Flash & \shortstack{MACs/\\frame} & \shortstack{Peak activation\\RAM} &
    \shortstack{PCK@0.10} & \shortstack{p95\\Latency} & \shortstack{HAR Top-1} & \shortstack{Fine-arm\\Recall} &
    \shortstack{PCK@0.10} & \shortstack{p95\\Latency} & \shortstack{HAR Top-1} & \shortstack{Fine-arm\\Recall} \\
    \midrule
    \multicolumn{13}{@{}l}{\emph{MCU-scale methods}} \\

    Tiny MobileNet Pose & 307.4K & 313.88~KiB & 13.39M & 197.05~KiB
    & 66.40\% & 0.061~s & $54.70{\pm}0.85$\% & 38.00\%
    & 62.22\% & 0.067~s & $81.37{\pm}1.19$\% & 83.56\% \\
    Uniform-Expansion Backbone & 435.6K & 439.96~KiB & 41.27M & 262.75~KiB
    & 72.90\% & 0.121~s & $68.55{\pm}0.30$\% & 51.80\%
    & 68.81\% & 0.127~s & $85.80{\pm}0.59$\% & 85.28\% \\
    Progressive-Expansion Backbone & 284.9K & 290.94~KiB & 25.14M & 262.75~KiB
    & 73.70\% & 0.076~s & $69.43{\pm}0.34$\% & 52.70\%
    & 64.40\% & 0.082~s & $84.05{\pm}0.27$\% & 85.24\% \\
    XiNet Pose & 607.5K & 603.46~KiB & 75.09M & 283.63~KiB
    & 72.08\% & 0.190~s & $60.45{\pm}0.80$\% & 41.79\%
    & 69.01\% & 0.209~s & $86.70{\pm}1.06$\% & 87.40\% \\
    Activation-Aware Backbone & 284.2K & 289.18~KiB & 29.54M & 248.64~KiB
    & 78.00\% & 0.107~s & $73.65{\pm}0.91$\% & 60.40\%
    & 75.49\% & 0.097~s & $85.52{\pm}0.85$\% & 84.47\% \\
    \textbf{\sysname} & 294.3K & 289.18~KiB & 29.56M & 248.64~KiB
    & \textbf{78.40\%} & 0.107~s & $\mathbf{74.99{\pm}1.06}$\% & \textbf{63.10\%}
    & \textbf{77.18\%} & 0.097~s & $\mathbf{87.46{\pm}0.12}$\% & \textbf{87.44\%} \\
    \midrule
    \multicolumn{13}{@{}l}{\emph{Non-MCU accuracy references}} \\
    YOLOv8n-Pose & 3.30M & -- & 4,648M & --
    & 84.40\% & -- & 89.00\% & 81.80\%
    & 88.2\% & -- & 92.00\% & 88.4\% \\
    YOLO11n Detector + RTMPose-S & 8.34M & -- & 4,447M & --
    & 85.60\% & -- & 90.30\% & 83.90\%
    & 93.5\% & -- & 95.11\% & 91.2\% \\
    \bottomrule
  \end{tabularx}
  \endgroup
\end{table*}

\subsubsection{Pose and Downstream Utility}
\sysname achieves the strongest pose and HAR performance among the MCU-scale methods on both datasets. On NTU RGB+D, it reaches 78.40\% PCK@0.10, $74.99{\pm}1.06$\% HAR Top-1, and 63.10\% fine-arm recall. The same ranking largely persists on Penn Action, where \sysname again leads the MCU-scale methods in all three metrics. The consistency across datasets suggests that the gain is not tied to a particular action distribution or capture setting. The baseline ordering also reveals that frame-level localization accuracy alone does not fully determine downstream skeleton utility. On Penn Action, XiNet Pose and Uniform Expansion achieve HAR accuracy comparable to the activation-aware backbone despite substantially lower PCK. HAR operates on temporal skeleton sequences, so its performance depends on the structure of the remaining pose errors, including which joints are affected and whether those errors persist coherently over time. Errors on joints carrying little discriminative motion may have limited impact on activity recognition, while smaller errors on wrists and elbows can alter the temporal signature of arm-centric actions. This distinction is reflected by fine-arm recall: Relative to the backbone-only variant, \sysname improves this metric more strongly than aggregate PCK on NTU, indicating that its gains are concentrated in joints that matter disproportionately for fine-grained motion recognition. The non-MCU references remain more accurate, quantifying the headroom available when the MCU resource envelope is relaxed.

\subsubsection{Resource Efficiency}
The comparison also shows that resource use alone is a poor predictor of pose quality. Relative to XiNet Pose, \sysname uses 51.6\% fewer parameters, 52.1\% less Flash, 60.6\% fewer MACs/frame, and 12.3\% less peak activation RAM, yet achieves higher PCK and HAR on both datasets. Its 289.18~KiB Flash footprint is also the smallest among the MCU-scale methods in Table~\ref{tab:end-to-end-comparison}. Uniform Expansion similarly uses 41.27M MACs/frame and 262.75~KiB peak activation RAM while remaining below \sysname in pose and downstream utility. At the other extreme, Tiny MobileNet has the lowest computation and peak activation RAM but loses substantial utility, especially on NTU. These operating points show that efficiency depends on how capacity is allocated, not only on the total resource budget.

Peak activation RAM clarifies this resource-accuracy trade-off. \sysname peaks at 248.64~KiB, below XiNet Pose at 283.63~KiB and Uniform and Progressive Expansion at 262.75~KiB. The activation-aware backbone has the same 248.64~KiB peak, showing that the distal-joint refiner improves pose and HAR utility without raising the measured maximum. Tiny MobileNet has a lower 197.05~KiB peak but is substantially less accurate. \sysname combines a controlled activation peak with moderate computation and the strongest MCU-scale pose and HAR utility. \S~\ref{sec:eval-allocation} examines this allocation effect directly.

\subsubsection{INT8 Robustness}
Table~\ref{tab:int8-retention} shows that \sysname retains its accuracy after INT8 conversion, reaching 77.95\% PCK@0.10 on NTU RGB+D and 77.14\% on Penn Action. Most methods lose less than two points, while Progressive Expansion drops sharply on NTU, suggesting greater sensitivity to activation quantization from heterogeneous intermediate ranges and repeated requantization. The backbone alone and \sysname show much smaller degradation, indicating that their accuracy advantage survives the INT8 deployment.

\begin{table}[t]
  \caption{FP32-to-INT8 PCK@0.10 retention.}
  \label{tab:int8-retention}
  \centering
  \scriptsize
  \begingroup
  \renewcommand{\arraystretch}{1.05}
  \setlength{\tabcolsep}{1.6pt}
  \begin{tabularx}{\columnwidth}{@{}>{\raggedright\arraybackslash}Xrrrrrr@{}}
    \toprule
    \multirow{2}{*}{Method} & \multicolumn{3}{c}{NTU RGB+D} & \multicolumn{3}{c}{Penn Action} \\
    \cmidrule(lr){2-4}\cmidrule(l){5-7}
    & FP32 & INT8 & $\Delta$ & FP32 & INT8 & $\Delta$ \\
    \midrule
    Tiny MobileNet Pose & 66.40\% & 65.87\% & $-0.53$ & 62.22\% & 62.01\% & $-0.21$ \\
    Uniform-Expansion Backbone & 72.90\% & 71.84\% & $-1.06$ & 68.81\% & 68.65\% & $-0.16$ \\
    Progressive-Expansion Backbone & 73.70\% & 64.20\% & $-9.50$ & 64.40\% & 64.19\% & $-0.21$ \\
    XiNet Pose & 72.08\% & 70.45\% & $-1.63$ & 69.01\% & 68.85\% & $-0.16$ \\
    Activation-Aware Backbone & 78.00\% & 77.45\% & $-0.55$ & 75.49\% & 75.33\% & $-0.16$ \\
    \textbf{\sysname} & \textbf{78.40\%} & \textbf{77.95\%} & $\mathbf{-0.45}$ & \textbf{77.18\%} & \textbf{77.14\%} & $\mathbf{-0.04}$ \\
    \bottomrule
  \end{tabularx}
  \endgroup
\end{table}

\begin{figure*}[t]
  \centering
  \begingroup
  \setlength{\unitlength}{0.96\textwidth}
  \begin{picture}(1,0.258)
    \put(0,0){\includegraphics[width=\unitlength]{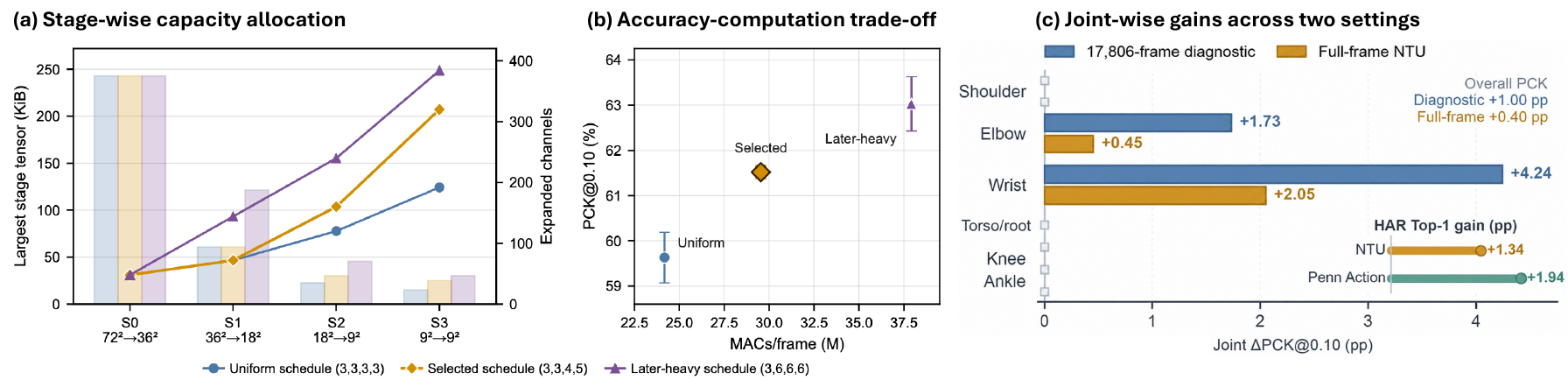}}
  \end{picture}
  \endgroup
  \captionsetup{labelfont=bf,textfont=normalfont}
  \caption{\textbf{Design analysis of the activation-aware capacity allocation and hierarchical distal-joint refinement.}
  (a) Stage-wise capacity allocation under a fixed largest individual activation tensor.
  (b) Accuracy--computation trade-off across the three controlled expansion schedules.
  (c) Joint-wise gains from the refiner, showing improvements concentrated on the elbow and wrist.}
  \label{fig:design-analysis}
\end{figure*}

\subsection{Design Analysis}
\label{sec:eval-design}
The preceding results established the operating point selected for the final model. We next examined how the activation-aware backbone distributed capacity under the constraint on the largest individual activation tensor and how distal-joint refiner contributed to the observed gains.
\subsubsection{Expansion Allocation}
\label{sec:eval-allocation}
\fig~\ref{fig:design-analysis}(a) shows why the selected expansion schedule assigned additional capacity to later stages. Across all three controlled schedules, the largest activation tensor occurred in Stage~0, at $1\times48\times72\times72$. In every schedule, the largest individual activation tensor was 248{,}832~B. After spatial downsampling, later stages retained substantial channel headroom while their activation tensors remained below this fixed tensor-size ceiling. The selected $(3, 3, 4, 5)$ schedule used this headroom to allocate most additional capacity at lower spatial resolutions.
This largest-tensor constraint is distinct from the canonical model's \finalPeakActivationKiB peak activation memory reported elsewhere.

\fig~\ref{fig:design-analysis}(b) quantifies the resulting trade-off between PCK and computation. Relative to Peak-Matched Uniform, the selected schedule increased PCK by 1.89 points at a cost of 5.37M additional MACs/frame. This corresponds to 0.352 PCK points per additional 1M MACs/frame. The aggressive later-heavy schedule increased PCK by a further 1.51 points but required 8.39M additional MACs/frame, reducing marginal efficiency to 0.180 PCK points per 1M MACs/frame. Marginal efficiency therefore declined by 48.9\%, indicating diminishing returns from further widening and supporting selection of the $(3,3,4,5)$ schedule.

\subsubsection{Targeted Distal Refinement}
\label{sec:eval-distal}
\fig~\ref{fig:design-analysis}(c) shows that the improvements from hierarchical distal-joint refinement were concentrated at the intended distal arm joints. On the 17{,}806-frame diagnostic evaluation used for module selection, wrist and elbow PCK increased by 4.24 and 1.73 points, respectively. PCK at the shoulder, torso/root, knee, and ankle remained unchanged at the reported precision. The full-frame NTU evaluation underlying Table~\ref{tab:end-to-end-comparison} showed the same joint-specific pattern. Wrist and elbow PCK increased by 2.05 and 0.45 points, respectively, while the other joints remained unchanged at the reported precision. Although overall NTU PCK increased by 0.40 points, HAR Top-1 accuracy and fine-arm recall increased by 1.34 and 2.7 points, respectively. A similar downstream pattern was observed on Penn Action. The concentration of gains at the wrist and elbow was consistent with the downstream improvements in arm-centric activity recognition. The refiner added only 10.1K parameters and 0.02M MACs/frame while preserving peak activation memory at \finalPeakActivationBytes.

\subsection{Live Deployment}
\label{sec:eval-live}

We next evaluate \sysname in three indoor deployment environments, Envs.~A, D, and E (\fig~\ref{fig:hardware-platform}), using the live-camera setup and the one-shot calibration procedure in \S~\ref{sec:calibration}. Across these environments, we evaluate pose accuracy and downstream HAR under deployment shift, followed by MCU power and adaptive execution. Finally, we study a separate long-horizon single-elderly monitoring deployment, a compelling application that requires continuous activity awareness while limiting exposure of raw visual data.

\subsubsection{Deployment Performance}
\begin{figure}[t]
  \centering
  \includegraphics[width=0.85\columnwidth]{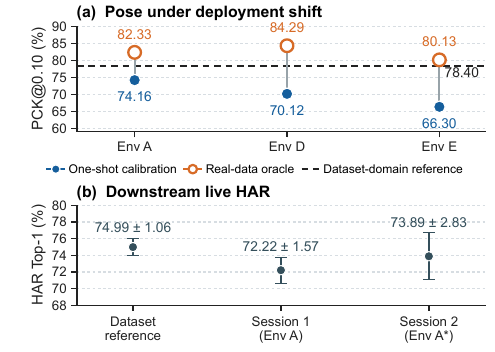}
  \captionsetup{labelfont=bf,textfont=normalfont}
  \caption{\textbf{Live deployment results.} (a) PCK@0.10 after one-shot site calibration. Orange and blue markers denote target-supervised references and one-shot calibration results. The dashed line denotes the dataset-domain reference. (b) Three-seed mean HAR Top-1 for the dataset reference and two independently collected Env~A sessions; Env~A* changes appearance and lighting.}
  \label{fig:live-deployment-results}
\end{figure}

Figure~\ref{fig:live-deployment-results}(a) evaluates pose-estimation accuracy after one-shot calibration to each deployment site. The dataset reference achieved 78.40\% PCK@0.10, whereas empty-scene calibration yielded 74.16\%, 70.12\%, and 66.30\% in Envs.~A, D, and E, respectively. Env.~A recovered most of the dataset-reference accuracy. The lower performance in Envs.~D and E likely reflects their more challenging, visually complex backgrounds, which were not fully captured by one-shot empty-scene calibration. With labeled target-person data, adaptation increased PCK to 82.33\%, 84.29\%, and 80.13\% in the three environments. These values exceeded empty-scene calibration by 8.17, 14.17, and 13.83 points, respectively, defining the improvement available from stronger target supervision.
Figure~\ref{fig:live-deployment-results}(b) tests whether the calibrated skeletons retained sufficient information for downstream human activity recognition. Live RGB frames were processed by the calibrated, frozen \sysname estimator. The resulting skeletons were encoded using the same 64-frame representation as in the dataset experiments. Session~1 achieved $72.22{\pm}1.57$\% HAR Top-1 in Env.~A, whereas independently collected Session~2 achieved $73.89{\pm}2.83$\% under different appearance and lighting conditions. These results were 2.77 and 1.10 points below the $74.99{\pm}1.06$\% dataset reference, respectively. Accordingly, the evaluation assessed skeleton-representation utility across capture conditions rather than cross-session transfer of a fixed HAR model.

\subsubsection{Power and Adaptive Execution}
We first characterize the MCU-side cost of continuous pose execution. MCU-side current is measured through the NUCLEO JP4/IDD path shown in \fig~\ref{fig:hardware-platform}. We measure the production firmware across three 60-s runs after a two-minute warm-up. At 3.293~V, the MCU draws $185.65 \pm 1.09$~mA, corresponding to $611.35 \pm 3.59$~mW. At an output rate of $8.311 \pm 0.067$ skeletons/s, this yields $73.57 \pm 0.84$~mJ per emitted skeleton. These values characterize the MCU-side cost while the full pose pipeline is continuously active.

\begin{figure}[t]
  \centering
    \captionsetup{labelfont=bf,textfont=normalfont}
  \includegraphics[width=0.85\columnwidth]{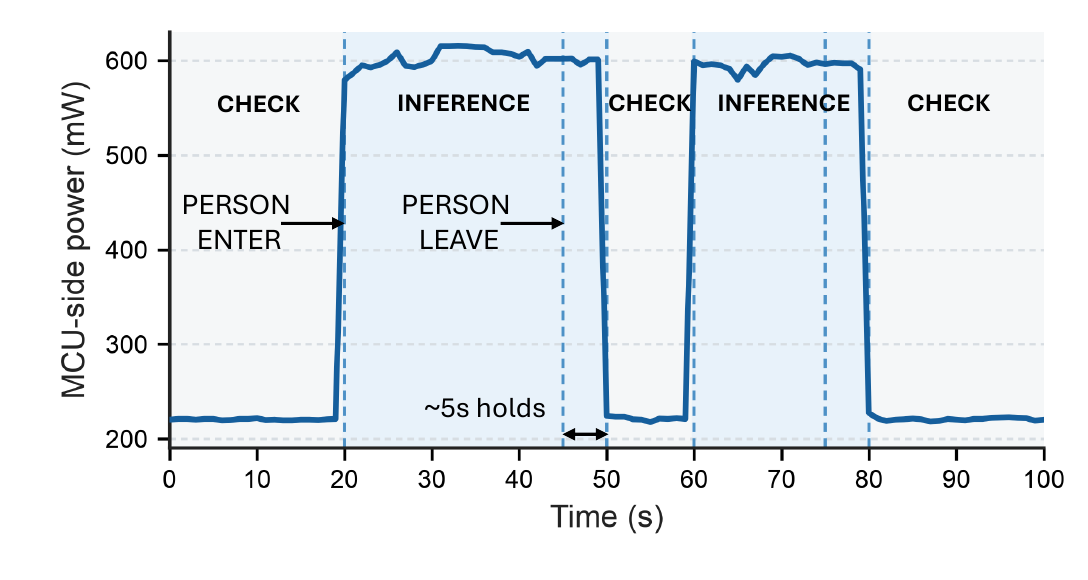}
  \caption{\textbf{Motion-triggered adaptive execution.} A 100-s power trace with two human-entry periods. \texttt{MOTION\_CHECK} performs one motion check per second and switches to full-rate \texttt{POSE\_INFERENCE} upon detecting motion, returning to \texttt{MOTION\_CHECK} after approximately 5~s of inactivity.}
  \label{fig:adaptive-power}
\end{figure}

To reduce this cost during inactive periods, adaptive execution switches between \texttt{MOTION\_CHECK} and \texttt{POSE\_INFERENCE} states. \texttt{MOTION\_CHECK} performs a frame-difference motion check once per second, requiring 743~$\mu$s per check and 576~B of additional state. Detected motion activates the full \sysname pipeline, while approximately 5~s of inactivity returns the system to \texttt{MOTION\_CHECK}. Motion detection to the first emitted skeleton takes 121~ms, and \texttt{POSE\_INFERENCE} sustains 8.222~FPS compared with 8.236~FPS for continuous execution. \fig~\ref{fig:adaptive-power} shows the resulting current trace across two human-entry periods. \texttt{MOTION\_CHECK} draws approximately 66--68~mA, compared with approximately 179--182~mA during \texttt{POSE\_INFERENCE} execution, reducing MCU-side current by about 63\% while the scene is inactive. The rapid transition back to full-rate inference shows that this saving comes with little loss in responsiveness or active throughput.

\subsubsection{Three-day Monitoring Case Study.}
\begin{figure*}[t]
  \centering
  \captionsetup{labelfont=bf,textfont=normalfont}
  \includegraphics[width=0.9\textwidth]{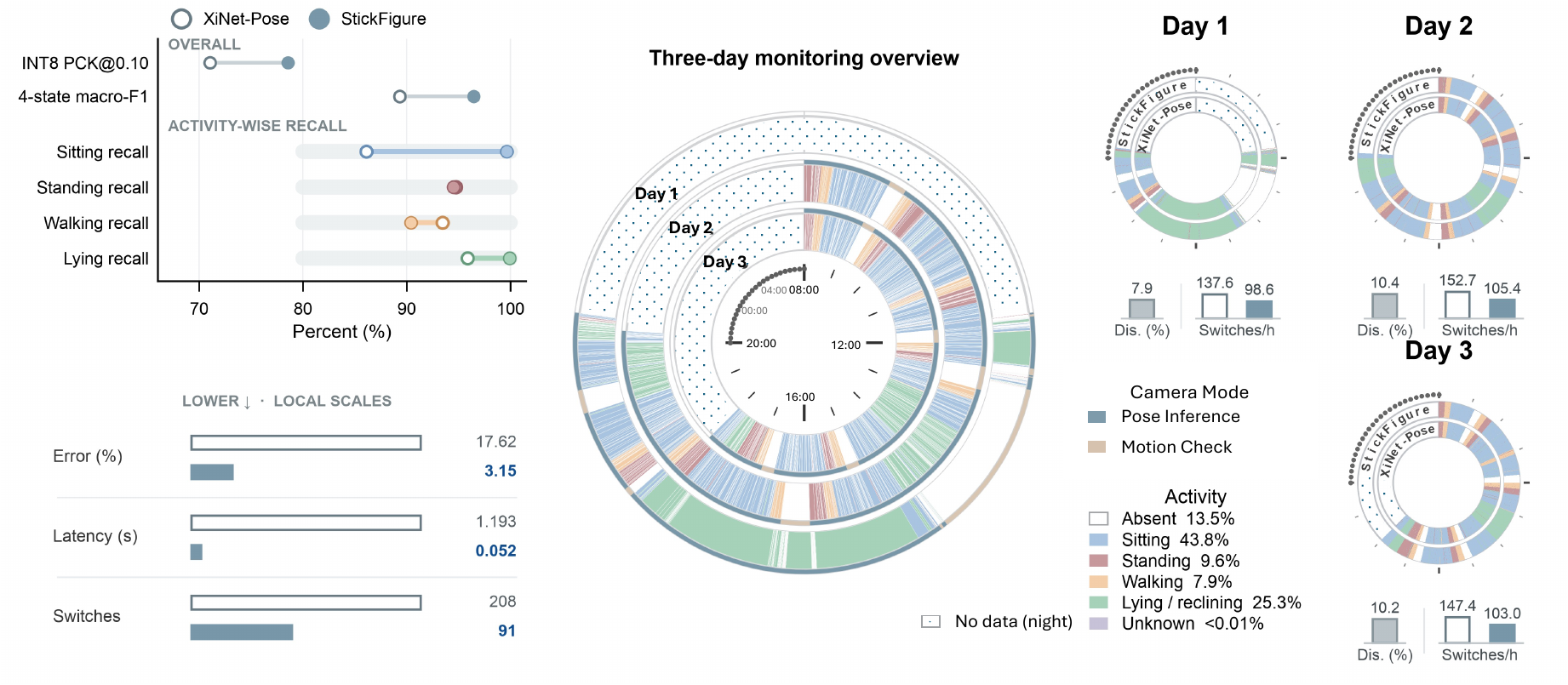}
  \caption{\textbf{Long-horizon monitoring case study.} \textbf{Left:} pose, HAR, and temporal metrics from a separate approximately 40-minute annotated session. \textbf{Center:} three daytime monitoring timelines. \textbf{Right:} day-specific prediction timelines, inter-method disagreement (Dis.), and activity-state switches per hour for the two estimators.}
  \label{fig:case-study}
\end{figure*}

We finally evaluate the complete pipeline in a fixed single-person daytime deployment involving an elderly participant. \fig~\ref{fig:case-study} summarizes three daytime sessions and a separate approximately 40-minute annotated segment. The three-day deployment has no dense ground truth, so we use it to characterize long-term prediction behavior under identical monitoring conditions for \sysname and XiNet Pose. Their predicted activity states disagree for 7.9\% to 10.4\% of the monitored intervals across the three days. \sysname also produces fewer activity-state transitions per hour on each day, with 98.6 versus 137.6, 105.4 versus 152.7, and 103.0 versus 147.4 transitions. These measurements indicate a more temporally stable activity stream, while the absence of ground truth prevents interpreting them as absolute recognition accuracy.

We therefore conduct a separate approximately 40-minute session with ground-truth annotations. \sysname achieves a four-state macro-F1 of 96.45\%, compared with 89.31\% for XiNet Pose. It also reduces mean absolute activity-duration error from 17.62\% to 3.15\%, median first-correct-frame delay from 1.193~s to 0.052~s, and erroneous state switches from 208 to 91. First-correct-frame delay is measured from the annotated state-transition boundary to the first skeleton predicted as the new state. These results show the more stable long-horizon activity stream is accompanied by improved state accuracy, duration estimation, and transition responsiveness.

\subsection{Cross-Platform Scaling}
\label{sec:eval-cross-platform}
We evaluate pose and HAR utility across four resource tiers represented by ESP32-S3, STM32H753, Raspberry Pi~4, and Jetson Orin Nano.
These open, programmable platforms provide reproducible proxies for MCU-class, low-cost Linux SBC, and accelerated embedded-vision hardware~\cite{espressif_esp32s3,stm32h753,raspberrypi4,nvidia_orin_nano}.
We discard candidates that violate INT8 execution, memory, storage, or the real-time gate of p95 inference latency \(\leq125\) ms and sustained throughput \(\geq8\) FPS.
 On ESP32-S3, we begin with a predefined \(5\times5\) width–resolution grid and extend the search when no candidate is feasible.
  After static deployment checks, ordered on-device probing identifies candidates near the feasible frontier.
  Candidates near the feasible frontier undergo formal timing with 20 warm-up runs followed by 200 pose inferences.
  Timing begins with the quantized input resident in memory and ends when the final 12-joint pose is available, excluding initialization and I/O.
  Among feasible candidates, we select each architecture’s highest-PCK@0.10 configuration using a one-epoch INT8 validation screen, then train it for eight epochs for final evaluation.
  Neither test-set pose accuracy nor HAR results are used during configuration selection.

\begin{figure}[t]
  \centering
  \captionsetup{labelfont=bf,textfont=normalfont}
  \includegraphics[width=0.95\columnwidth]{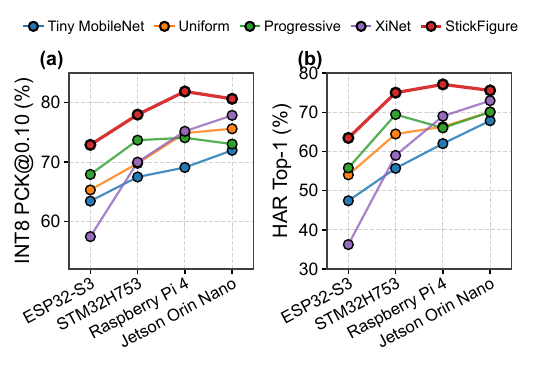}
  \caption{\textbf{Cross-platform INT8 pose and downstream utility.} (a) INT8 PCK@0.10 and (b) downstream HAR Top-1 for the best feasible operating point of each architecture family on ESP32-S3, STM32H753, Raspberry Pi~4, and Jetson Orin Nano. Each point is the highest-performing configuration satisfying the cross-platform real-time gate of p95 inference latency $\leq125$~ms and sustained throughput $\geq8$~FPS. \sysname achieves the highest pose and HAR utility across all four platforms.}
  \label{fig:cross-platform-scaling}
\end{figure}

\fig~\ref{fig:cross-platform-scaling} shows that \sysname retains more pose and HAR utility as resource budgets tighten.
Performance differences are small on larger platforms but widen under tighter budgets, where competing architectures require more aggressive scaling.
At the ESP32-S3 endpoint, \sysname achieves 72.9\% INT8 PCK@0.10 and 63.5\% HAR Top-1, exceeding the next strongest method by 5.0 and 7.7 percentage points, respectively. \sysname therefore provides the highest-utility feasible operating point across all four tiers and degrades more gradually under resource-driven scaling.

\section{Related Work}
\label{sec:related}

\textbf{Privacy-\allowbreak conscious visual monitoring.}
Privacy-\allowbreak conscious camera and visual-sensing systems intervene throughout the visual-data path. Access control, trusted execution, private query release, and mediated interfaces restrict how captured video is stored, processed, or disclosed~\cite{beugin2022smartcamera,poddar2020visor,cangialosi2022privid,jana2013darkly,aditya2016ipic}. Other work transforms optics, reduces spatial resolution or motion fidelity, or controls the physical capture state~\cite{hinojosa2021privacyoptics,hinojosa2022privhar,ryoo2017lowresolution,kumawat2022motion,shalawadi2024privacycovers}. Although these mechanisms assume different trust boundaries, many retain appearance-rich RGB within a trusted domain. Other systems change the form of data leaving the camera. WISPCam, BackCam, and NeuriCam reduce transmitted data through selection, compression, or downstream reconstruction, but retain image-bearing outputs~\cite{naderiparizi2015wispcam,naderiparizi2016wispcam,josephson2019backcam,veluri2023neuricam}. Battery-free semantic cameras and HyperCam perform local inference and export task-specific classifications or detections~\cite{giordano2020batteryfree,lee2025hypercam}. Skeleton-based recognition instead treats joint sequences as structured descriptions of articulated body motion. ST-GCN and CTR-GCN operate directly on trajectories encoding body geometry and temporal dynamics~\cite{yan2018stgcn,chen2021ctrgcn}. Compared with RGB, skeletons omit much of the appearance and background information without providing a formal privacy guarantee~\cite{shi2019directedgnn,zhou2023discriminative}. Bultmann and Behnke demonstrated 2D pose export from smart edge sensors for backend multi-view fusion on more capable embedded hardware~\cite{bultmann2021smartedge}. \sysname studies whether continuous articulated pose extraction can be sustained on a Cortex-M-class camera under tight memory and runtime constraints while retaining downstream temporal utility.

\noindent\textbf{Pose extraction under MCU constraints.}
Exporting a skeleton stream requires estimating pose from RGB on the camera. Accuracy-oriented estimators such as SimpleBaseline and HRNet use deconvolutional heatmaps or persistent high-resolution features, which incur large intermediate activations~\cite{xiao2018simplepose,sun2019hrnet}. Lite-HRNet, LitePose, and RTMPose reduce this cost through lightweight high-resolution processing, streamlined branches, and deployment-oriented coordinate prediction~\cite{yu2021litehrnet,wang2022litepose,jiang2023rtmpose}. These efficient models primarily target smartphones, mobile CPUs/GPUs, single-board computers, or accelerators. XiNet-Pose is the closest Cortex-M precedent, demonstrating pose estimation on an STM32H7-class MCU~\cite{ancilotto2024xinetpose}. In this regime, parameters and MACs alone do not determine deployability. Peak activation memory constrains model activations, while camera and runtime buffers, workspace, and firmware state contribute separately to deployment-level memory. Quantized operator support and kernel behavior determine memory use and latency; joint-localization quality determines whether the skeleton remains useful. CMSIS-NN and TensorFlow Lite Micro supply quantized kernels and embedded runtimes, while MCUNet and MCUNetV2 demonstrate architecture--runtime co-design and activation-aware execution~\cite{lai2018cmsisnn,david2021tflm,lin2020mcunet,lin2021mcunetv2}. The unresolved operating point is continuous articulated pose extraction on a Cortex-M-class camera within tight memory and runtime limits while retaining downstream temporal utility. \sysname targets this regime by converting local RGB into a skeleton stream and explicitly evaluating its downstream HAR utility.

\section{Discussion and Future Work}

\noindent\textbf{An enforceable camera-side privacy boundary.}
We place the trusted boundary at the camera MCU, the earliest programmable stage after capture, thereby confining raw RGB to the device and minimizing the trusted computing base. Our choice of a compact, fixed-schema pose interface further makes it possible to enforce this boundary quantitatively at the device egress. \sysname exports \sfPacketBytes SFSK records at 8.222~FPS, yielding a nominal payload rate of approximately 5.0~kb/s. By comparison, a single \(144\times144\) RGB565 frame contains 41,472~B, more than 500 times as much data as one SFSK record. Because legitimate traffic is bounded in both format and rate, future implementations could admit only valid SFSK records and cap transmission near the expected skeleton-stream rate. The skeleton stream would remain within this budget, whereas raw frames and substantially larger intermediate representations containing sensitive information would exceed it by orders of magnitude and could not pass through the permitted interface.

\noindent\textbf{Calibration with less scene exposure.}
\sysname adapts to a new site through empty-scene calibration, allowing users to control what is visible during setup, but the image can reveal room appearance. As a simple mitigation, we tested reducing the calibration input before adaptation. Strong blur largely preserves calibration accuracy, while downsampling to \(36\times36\) produces comparable or better PCK across the three deployment sites; a more aggressive \(18\times18\) reduction is less consistent. This suggests that calibration does not require a high-fidelity view of the scene, and deployments can downsample the snapshot before calibration to reduce exposed visual detail. Future systems could perform this reduction on-device and immediately discard the source image.

\noindent\textbf{Multi-person monitoring.}
The current design intentionally targets single-person or primarily single-person monitoring, matching private-room, assisted-living, and single-elderly settings. A natural extension is lightweight multi-person pose extraction with temporal association and multiple skeleton streams. Future systems could allocate computation dynamically among active subjects while bounding MCU memory and latency as occupancy grows, making multi-person support a resource-allocation problem as well as a perception and tracking task.

\noindent\textbf{Configurable semantic interfaces.}
The fixed 12-joint stream suits the body-motion tasks studied here, but other applications may require different semantic granularity. Future versions could configure joint subsets, coordinate precision, temporal sampling rates, confidence fields, or narrowly scoped additional semantic signals. Presence monitoring, for example, may require substantially less information than arm-motion analysis. Such configurability could balance downstream utility, communication and computation cost, and information exposure without reverting to RGB transmission.

\section{Conclusion}
\label{sec:conclusion}
\sysname shows that a privacy-conscious camera can keep RGB local during monitoring and export only pose streams for downstream body-motion sensing. Its design jointly considers pose extraction, MCU resource constraints, and temporal utility: the activation-aware backbone limits large high-resolution intermediate activation tensors through late channel expansion, while the hierarchical distal-joint refiner improves task-critical distal-joint localization. Pose-source-matched HAR evaluates whether the resulting streams remain useful beyond frame-level pose accuracy. We evaluate \sysname across public benchmarks, live deployments, and cross-platform experiments. The results show that \sysname delivers strong pose and HAR utility under MCU constraints, sustains real-time execution on an STM32H753 MCU, and degrades more gracefully than SOTA baselines as computation resources tighten.
We believe \sysname sheds light on ubiquitous vision intelligence on massive widely deployed IoT cameras.